# SHDM-NET: Heat Map Detail Guidance with Image Matting for Industrial Weld Semantic Segmentation Network

Qi Wang and Jingwu Mei

*Abstract*—In actual industrial production, the assessment of the steel plate welding effect is an important task, and the segmentation of the weld section is the basis of the assessment. This paper proposes an industrial weld segmentation network based on a deep learning semantic segmentation algorithm fused with heatmap detail guidance and Image Matting to solve the automatic segmentation problem of weld regions. In the existing semantic segmentation networks, the boundary information can be preserved by fusing the features of both high-level and low-level layers. However, this method can lead to insufficient expression of the spatial information in the low-level layer, resulting in inaccurate segmentation boundary positioning. We propose a detailed guidance module based on heatmaps to fully express the segmented region boundary information in the low-level network to address this problem. Specifically, the expression of boundary information can be enhanced by adding a detailed branch to predict segmented boundary and then matching it with the boundary heat map generated by mask labels to calculate the mean square error loss. In addition, although deep learning has achieved great success in the field of semantic segmentation, the precision of the segmentation boundary region is not high due to the loss of detailed information caused by the classical segmentation network in the process of encoding and decoding process. This paper introduces a matting algorithm to calibrate the boundary of the segmentation region of the semantic segmentation network to solve this problem. Through many experiments on industrial weld data sets, the effectiveness of our method is demonstrated, and the MIOU reaches 97.93%. It is worth noting that this performance is comparable to human manual segmentation ( MIOU 97.96%).

## I. INTRODUCTION

AS an essential material forming process, welding is widely used in automobile manufacturing. The quality of the weld directly affects the strength of the bonded metal. Therefore, to ensure the quality and safety of the welding area, it is necessary to conduct a quality inspection the appearance and size of the welding area after welding. Therefore, to ensure the quality and safety of the welding area, it is necessary to conduct a quality inspection on the appearance and size of the welding area after welding. In order to carry out the quality inspection on the welding seam area, the welding seam area should be divided first. At present, the segmentation inspection of the welding seam area is mainly measured and evaluated by the hand-painted identification of professionals. This method consumes many human and material resources and depends on the tester's experience for segmentation quality, which is highly subjective. Therefore, an efficient and accurate weld image segmentation algorithm has become urgent for industrial sectors. Since the segmentation effect of the weld zone directly affects the results of welding quality detection, the accuracy of the segmentation algorithm is crucial. Traditional segmentation algorithms, such as threshold segmentation [1-3], edge detection segmentation [4, 5], watershed algorithm [6], can only be used to deal with simple scenes. As for the segmentation of the weld image, difficulties in welding zone segmentation are mainly incarnated in two aspects. Difficulties in welding zone segmentation are mainly incarnated in two aspects. On the one hand, On the boundary area of the weld image is ambiguous, and the weld and welding material is fusion with each other. On the other hand, the projective shape of the welding zone is susceptible to comprehensive factors such as shooting angle and welding technology. Uniform criterion is hard to establish with traditional segmentation method. They will lead to obvious problems such as the missing and error detection of edges in the traditional segmentation method. Fig.1 shows the schematic diagram of some metal sections.

Recently, DCNN(deep convolutional neural network)has been widely used in semantic segmentation. FCN [7], Seg Net [8], U-Net [9], and Deeplab V3 + [10] have achieved great success in semantic segmentation. However, most of these methods have poor segmentation performance for boundary areas. The main reason is that the detailed boundary information of the segmentation region will be lost along with the decrease of image resolution with the stacking of pooling and convolution layers. The loss of these spatial details will lead to poor network predictions. To make up for the lack of deep layer features in spatial details and boundary information, some research scholars adopt the Encoder-Decoder structure [9, 10, 11, 12]. By introducing a lateral connection fusing low-level and high-level layer features to increase spatial information characteristics and improve segmentation accuracy. In addition, Bisenetv1 [12] and Bisenetv2[13] proposed dual-stream paths for low-level details and high-level context information to harmonize spatial and semantic information. STDC[15] set up a single-path decoder, which uses a detailed information guide to learn low-level details. Although these methods improve the spatial information of DCNN deep layers features, they also bring much useless information, such as texture, shape and color information of background region, which cause interference to deep feature classification. In order to achieve a better welding seam segmentation effect, we propose a segmentation module loaded on DCNN guided by Gaussian heat map details and fused with the Matting algorithm. This module can be added to a DCNN-based semantic segmentation network to achieve increased segmentation accuracy and more precise segmentation edges. The splicing of the various parts of the model is shown in Fig. 2.

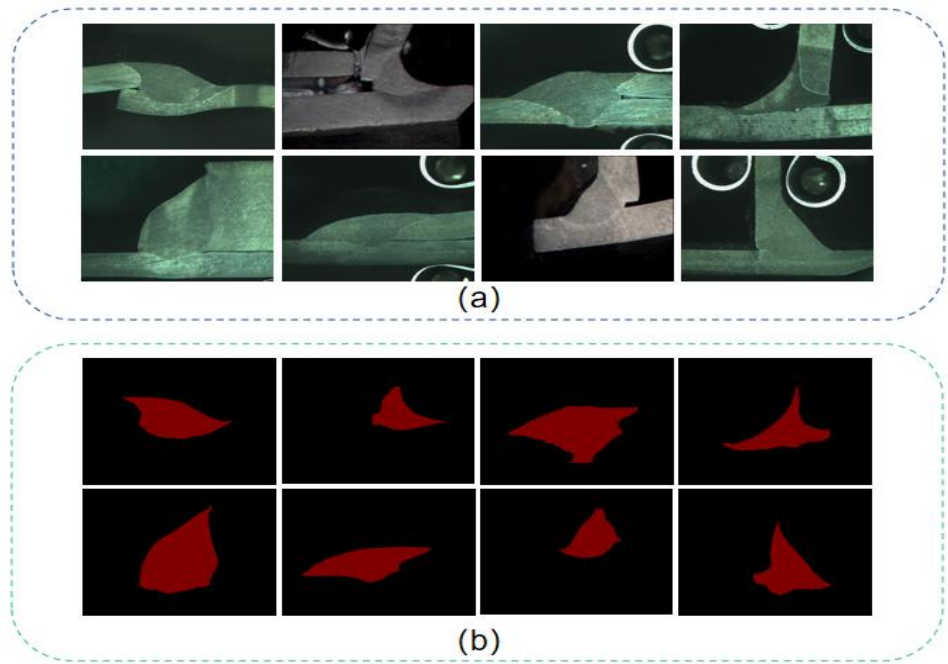

**Fig. 1**. Welding images and their segmentation ground-truth, (a) images of the welding area, (b) the actual label of the corresponding weld, (c) Enlarged image of the edge of the weld area, we can find the boundary area of the weld image is ambiguous, and the weld and welding material is fusion with each other

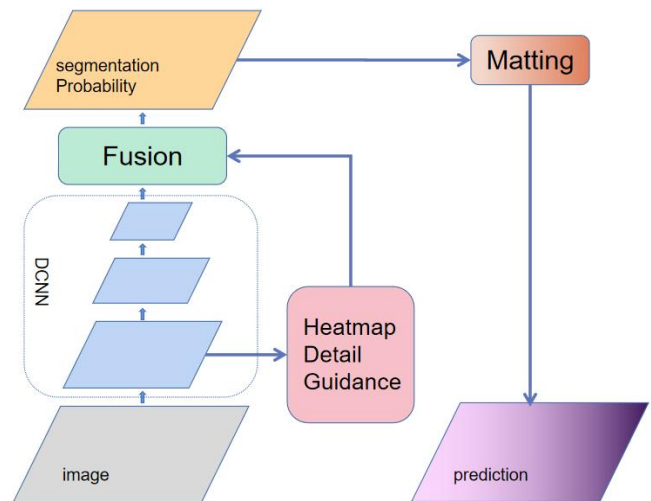

**Fig. 2.** Insertion of Gaussian heatmap guidance module and Image Matting algorithm in semantic segmentation network.

In the model training stage, Gaussian heatmap detail guidance is introduced into the low-level layer. We use the weld boundary heatmap generator to generate the Gaussian heatmap ground-truth of weld edges. And the detail generation head is introduced into the low-level network to generate the probability map of segmented edges. Then MSE (Mean Square Error) is used to optimize the learning task of segmented edge information, which makes low-level network pays more attention to the weld boundary information. Then the information is introduced into the high-level network fuse with the deep semantic information through the lateral connection. It is worth noting that we do not need to introduce this structure in prediction. It is only introduced in training. Therefore, the structure can improve segmentation accuracy without increasing the prediction time.

Popular semantic segmentation networks usually introduce a semantic segmentation head on a low-resolution feature map to obtain a probabilistic prediction map. The probability prediction map is then restored to the input image resolution by interpolation. This approach has the disadvantages of blurred segmentation results and unclear segmentation boundaries. We have designed a new semantic segmentation head to improve the above disadvantages.

The seam edge is significant for weld quality detection. However, popular semantic segmentation algorithms focus on pixel-by-pixel classification. Most methods don't pay enough attention to the connections of neighborhood pixels, where pixels with the same semantic meaning are classified into the same class, resulting in some class chunks, with no consideration given to whether the chunks are connected to each other. Thus, we introduce the closed-form Matting algorithm to solve this problem. Closed-form Matting algorithm uses a Trimap as a constraint to obtain the optimal solution of foreground alpha channel with an image-based local smooth hypothesis. Due to the local smooth hypothesis of Closed-form matting, the algorithm considers the local area pixel and texture changes. So the matting algorithm can refine the edge better under the condition of better Trimap identification. The implementation method of this module is as follows: firstly, the probability graph obtained from the semantic segmentation network is converted into a Trimap graph, and then the matting algorithm is introduced to refine the edge to get the final segmentation result. To demonstrate the effectiveness of the proposed model, we design a simple semantic segmentation network called SHDM-NET as shown in Fig. 3. We describe the structure of SHDM-NET in detail in section 3.

The main contributions of this paper can be summarized as follows:

•This paper proposes an efficient and accurate segmentation algorithm for industrial welds. This algorithm introduces the heatmap-based detail guidance module to guide the low-level layer to focus on the segmentation boundary information to save the location information of the segmentation edge more accurately, without additional computational costs in reasoning.

• A new semantic segmentation head structure is proposed, allowing for improved accuracy of semantic segmentation.

•To optimize the effect of semantic segmentation of the boundary regions, we introduce Closed-form Matting to refine the segmentation performance.

• The experimental results show that SHDM-NET far outperforms the other networks for the weld segmentation data set. Specifically, the introduction of Heatmap details guidance MIOU to reach 97.61%, and the MIOU reaches 97.93% with Closed-form Matting, which reaches the human manual segmentation.

## II. RELATED WORK

The existing weld segmentation algorithms can be divided into two categories. One is based on the traditional digital image processing method. The other is based on the depth learning method.

Classical algorithms

Among the methods based on classical digital image processing, Muthukumaran et al. [16] proposed an improved weld edge recognition method based on Otsu. Chen et al. [17] proposed a weld identification algorithm based on a distance diagram and understanding the prior knowledge of weld images. Doyoungchang et al. [18] used a Gaussian filter to eliminate noise from raw data, applied the differential feature point detection algorithm to filter data to detect the shape and position indicating the welding gap. Malarvel et al. [19] used the median filter to denoise the weld image, threshold segmentation for primary segmentation, morphological filter to eliminate excessive segmentation, and watershed algorithm to separate the internal region to achieve the purpose of segmentation of the weld region.

CNN for weld segmentation

In recent years, deep learning methods have been widely applied in image segmentation. However, through literature review, it is found that few deep learning methods have been developed for metal weld segmentation. WANG et al. [20] adopt convolutional neural networks and atrous convolutional spatial pyramids to segment the weld area. Han Yu et al. [21] proposed a welding seam segmentation network based on Resnet18 and ADSM (Adaptive Depth Selection Mechanism), and MIOU reached 86% on the self-built SRIF-CDS data set. Yi et al. [22]achieved good results in the field of Submerged Arc Welding (SAW) using a semantic segmentation network based on Fast-SCNN.

III. METHOD

SHDM-NET segment weld images from coarse to fine, including two steps: semantic segmentation and edge refine. For the semantic segmentation of weld seams, we designed a segmentation model based on Deeplabv3 +, which was guided by Gaussian heatmap details and fused with Matting algorithm to perform semantic segmentation and generate masks. Then, the closed-form Matting algorithm [23] is used to optimize the segmentation edge to improve the segmentation accuracy further.

**3.1 Segmentation Network Model**

3.1.1 Introduction to Network Structure

The final goal of weld seam segmentation is to obtain an accurate weld mask. The weld mask is a binary image of the same size as the input image. The pixel with a value of 1 represents the weld area, and the pixel with 0 represents other areas. This work designed an SHDM-NET to achieve this goal, as shown in Fig. 3. The network structure consists of two parts: encoder and decoder. In the encoder section, we use a pre-trained resnet50 network as the encoder's backbone, and the semantic information is extracted using the semantic feature extraction path of deeplabV3+. In layers 2, 3, and 4, feature maps are generated with downsampling ratios of 1/8, 1/16, and 1/32, respectively. The ASPP of deeplabV3+ is used, and its structure is shown in the ASPP module in Fig. 3. (a), which is visualized in detail in Fig. 4. Multi-scale parallel detection of feature maps is carried out using different scale DSDC (depthwise separable dilated convolution) while ensuring the lightweight of the network. The semantic information in the feature map can be extracted at different scales. In the segmentation network designed by us, ASPP is composed of a 1×1 convolutional layer, three 3×3 atrous convolutional layers with rates of 6,12,18, and a global average pooling layer.

The weld segmentation requires high accuracy in the edge region. Therefore, this paper up-samples the feature map four times. The feature map is turned to the same size as the input image. The prediction probability map is obtained by a semantic segmentation head. The weld segmentation head is composed of a set of convolution kernels with output dimension N followed by a softmax activation function. In weld segmentation, all the pixels should be classified as weld or not. $f(i,j) = 1$, if $(i,j)$ is in a weld region; $f(i,j) = 0$, if $(i,j)$ is not in a weld region

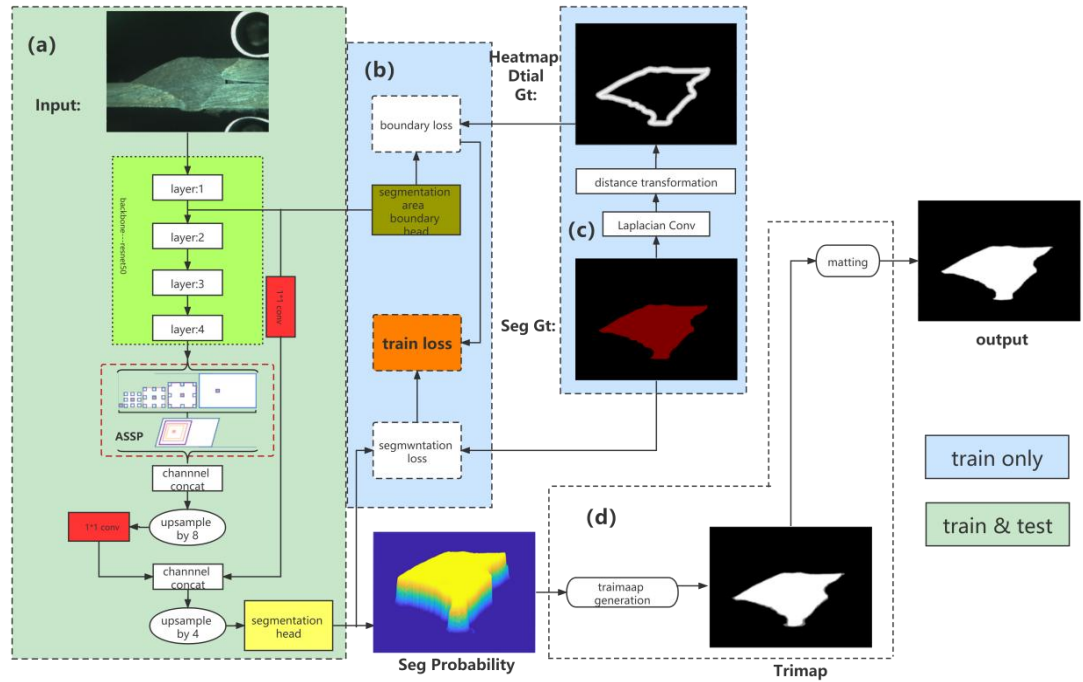

**Fig. 3.** SHDM-NET network structure diagram

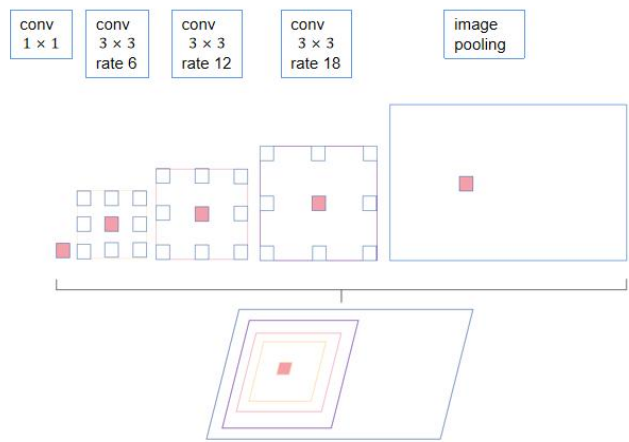

**Fig. 4.** ASPP

### 3.1.2 Gaussian heatmap guidance

In order to learn the low-level layer features better and pay attention to the boundary details of the weld area, we designed a segmentation edge details guidance module. First, as shown in Fig. 4., the edge is generated from the ground-truth of the welds through the Laplacian edge detection operator. Then, the distance between each pixel point in the figure and the edge of the weld is obtained by distance transformation. The ground-truth value of the Gaussian heatmap of the edge of the weld is generated by Gaussian function. As shown in Fig.3(a)., we connected a segmentation edge generator after Layer1 to generate the feature map of the weld edge. Specifically, a segmentation prediction task of pixel-by-pixel binary classification is performed here to predict the probability graph of the weld edge. Then, the boundary loss is calculated by the GT of the Gaussian heat map of the weld edge. This loss function can guide the prediction of weld edge probability maps. It makes low-level network learning pay more attention to the detailed features of the weld edge. Fig. 5. shows the visually interpretation of layer1 feature map. It validates the heatmap detail guidance module in SHDM-NET.

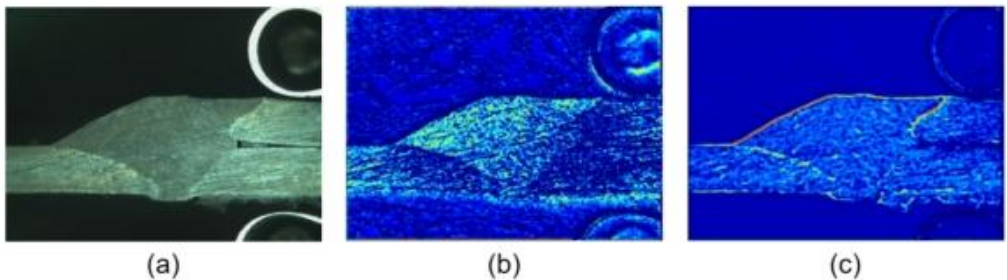

**Fig. 5.** Visual interpretation of layer1 feature map: (a) The original image (b) represents the result without heatmap detail guidance (c) represents the result with heatmap detail guidance. The visualization shows that after the detail guidance of heatmap, the network will pay more attention to the weld boundary area, edge, and corner points.

3.1.3 Gaussian heatmap generation

In order to generate heatmap detail GT, as shown in Fig 3(c), we first use the Laplacian edge detection operator to extract the boundary of the Seg-GT segmentation region. Since the pixels of the same segmentation area are labeled with the same values in Seg-GT, it is proper to extract the weld boundary by the Laplacian edge detector. The formula is shown in Eq(1) and Eq(2). The Boundary is calculated according to Eq(3).

$$\text{Laplacian kernel} = \begin{bmatrix} 1 & 1 & 1 \\ 1 & -8 & 1 \\ 1 & 1 & 1 \end{bmatrix} \qquad (1)$$

$$\text{R=Seg-GT*Laplacian kernel} \qquad (2)$$

$$B_{(x,y)} = \begin{cases} 1 & , R_{(x,y)} > 0.1 \\ 0 & , \qquad else \end{cases} \qquad (3)$$

Where * indicates convolution operation. R represents the result of Seg-Gt convoluted by Laplacian edge detector, $(x,y) \in R^{H \times W}$. B is the binary image of weld boundary which has the same size with Seg-Gt. A pixel is of boundary, if $B(x,y) = 1$. The boundary generating effect is shown in Fig 6 (c). Due to the extremely low proportion of boundary in the whole image, positive and negative samples are out of balance. In this case, boundary features are difficult to learn.

In order to better integrate boundary into feature learning, we convert boundary information into a boundary heatmap to help feature learning better. The response of each pixel in the boundary heatmap is determined by its distance to the nearest boundary pixel. The detailed definition of the boundary heat map is as follows. First, distance transformation is performed on the binary graph B of the weld boundary to obtain the boundary distance response graph D. The value of each point in the figure is the distance from the point to the nearest pixel on the boundary. We map D to the ground-truth boundary heat map M by Gaussian transformation and set distance threshold 3σ. This transformation makes the boundary heat map pay more attention to the boundary region.

$$M_i(x,y) = \begin{cases} \exp\left(-\frac{D(x,y)^2}{2\sigma^2}\right), & \text{if } D(x,y) < 3\sigma \\ 0, & \text{else} \end{cases} \quad (4)$$

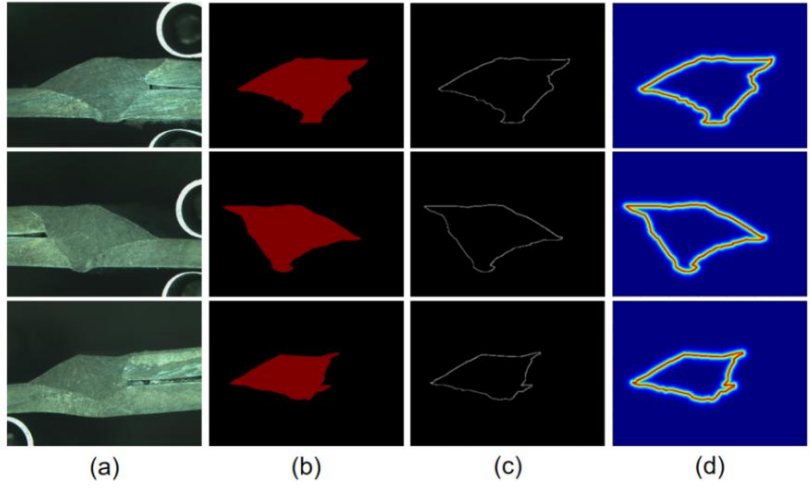

(a)　　　　(b)　　　　(c)　　　　(d)

**Fig. 6.** Generating process of boundary detail heatmap: (a) the original weld images (b) Seg-GT (c) boundary of weld seam (d) the heat map of weld seam boundary.

### 3.1.4 Loss function

As shown in Fig 3(a, b, c), the loss function consists of two parts: segmentation loss $l_{focal}$ and boundary loss Mse loss

$$\text{loss} = w_1 * l_{focal}(P_s, G_s) + w_2 * \text{Mse loss}(Pb, Hb) \quad (5)$$

Because segmentation loss $l_{focal}$ represents the quality of the final network segmentation output, network training should be more committed to reduce segmentation loss. In this paper, we set $w_1$=0.8, $w_2$=0.2. For segmentation loss use the Focal Loss function. To predict segmentation picture of height H and width W, the loss formula is as follows:

$$\text{segmentation loss} = l_{focal}(P_s, G_s) = -\alpha_t(1-p_t)^\gamma \log(p_t)$$

$$p_t = \begin{cases} P_s & G_s = 1 \\ 1 - P_s & \text{else} \end{cases} \quad (6)$$

Where, $p_s \in R^{H \times W}$ represents the predicted segmentation region, and $G_s \in R^{H \times W}$ represents the corresponding segmentation ground-truth value, $\alpha_t$=1, $\gamma$=0.2. For the case shown in Fig 4(b), we first use a segmented region boundary generation head to generate the boundary prediction map. The generation head contains a $3 \times 3$ Conv-BN-ReLU operator and a $1 \times 1$ convolution kernel to compress the number of feature map channels to 1. The boundary information is then learned by computing the boundary loss with the corresponding boundary detail heatmap GT. In this paper, we use the mean squared error loss as a measure of boundary prediction, and the mathematical expression for a boundary prediction map of height H and width W is as follows.

$$\text{boundary loss} = \text{Mse loss}(Pb, Hb) = \frac{\sum |Pb_i - Hb_i|^2}{H \times W} \quad (7)$$

where $Pb_i \in R^{H \times W}$ denotes the predicted boundary image, and $Hb_i \in R^{H \times W}$ represents the corresponding Gaussian heatmap boundary ground-truth value.

## 3.2 Matting algorithm edge optimization

After the original image is processed by our segmentation network, we can obtain a predictive map. The value of a pixel on the map indicates the probability that it belongs to weld seam area. The traditional semantic segmentation algorithm directly classifies the probability map into foreground and background by threshold. It generally leads to the following typical problems: (1) discontinuous and jagged boundaries, (2) holes, (3) discrete outliers. These shortcomings are unacceptable for weld quality inspection. To solve these problems, we use the Closed-form Matting algorithm to refine the segmentation

Let $I_i$ be the value of the $i$-th pixel in the image. It is weighted by the foreground point $F_i$ and background point $B_i$, ($i \in \Omega = \{1, \ldots, W \times H\}$) the mathematical expression is as follows:

$$I_i = \alpha_i F_i + (1 - \alpha_i) B_i \qquad (8)$$

where $\alpha_i (i \in \Omega)$ is the foreground transparency of the i-th pixel. All of $\alpha_i$ compose an alpha matrix with length W and width H. Therefore, when the Alpha matrix is determined, the original picture can be split into a foreground and background. The Closed-form Matting algorithm is based on the assumption of local smoothness of the image and the Trimap which is the user's rough annotation of the foreground, background, and fuzzy region. It deduces the form of the closed solution of the alpha matrix through the algebraic method. The foreground can be extracted according to the alpha matrix.

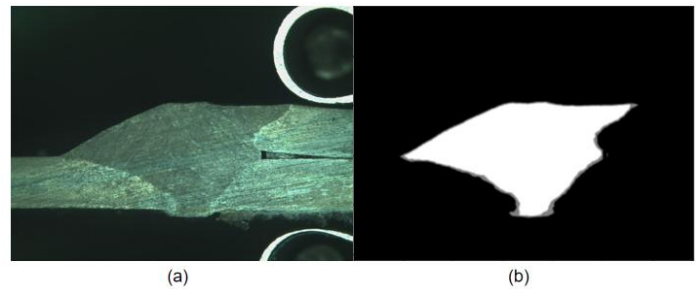

<center>(a)                    (b)</center>

**Fig.7.** (a) Original steel plate weld picture (b) The Trimap corresponding to the original image
(Trimap=1--->white,Trimap=0.5--->gray,Trimap=0--->black)

3.2.2Closed-form Matting solution

Assuming that the image is locally smooth, that is, in a small window w ($3 \times 3$), F and B are fixed, then equation (8) can be rewritten as:

$$\alpha_i \approx aI_i + b \qquad (10)$$

$$a = \frac{1}{F_i - B_i}, b = -\frac{B_i}{F_i - B_i}$$

If we can find αi, a and b making equation (9)being established on the window W, then equation (8) is solvable.

$$J(\alpha, a, b) = \sum_{j \in I} \left( \sum_{i \in w_j} (\alpha_i - a_j I_i - b_j)^2 + \epsilon a_j^2 \right) = \sum_k \| G_k \begin{bmatrix} a_k \\ b_k \end{bmatrix} - \bar{\alpha}_k \|^2 \qquad (11)$$

$$G_k = \begin{bmatrix} I_k & 1 \\ \sqrt{\varepsilon} & 0 \end{bmatrix} \qquad \bar{\alpha}_k = \begin{bmatrix} \alpha_k \\ 0 \end{bmatrix}$$

This is a typical least-squares problem. We can get answers by solving for $a_k, b_k$.

$$\begin{bmatrix} a_k \\ b_k \end{bmatrix} = (G_k^T G_K)^{-1} G_k^T \overline{\alpha_k} \qquad (12)$$

Thus, a and b can be eliminated from Equation(11) and the problem can be transformed into an optimization problem with only one variable, α.

$$J(\alpha) = \min_{a,b} J(\alpha, a, b) = \alpha^T L \alpha$$

$$L = \sum_{k|(i,j)\in w_k} \left( \delta_{ij} - \frac{1}{|w_k|} \left( 1 + \frac{1}{\frac{\epsilon}{|w_k|}+\sigma_k^2}(I_i - \mu_k)(I_j - \mu_k) \right) \right) \qquad (13)$$

$\delta_{ij}$ is equal to 1 only if i is equal to j, and 0 otherwise. $\mu_k$ and $\sigma_k^2$ are the mean and variance of pixels in the sub-window, respectively. The solution of an alpha matrix can be reduced to the following form under the Trimap constraint given by the user.

$$\alpha = \text{argmin } J(\alpha) = \text{argmin } \alpha^T L \alpha, \quad \text{s.t. } \alpha_i = s_i, \forall i \in S \qquad (14)$$

Where S is a non-gray area in the Trimap, $s_i$ is 1 in the white area is 1, and 0 in the balck area. Rewrite L and α as follows:

$$L = \begin{bmatrix} L_M & B \\ B^T & L_U \end{bmatrix} \quad \alpha = \begin{bmatrix} \alpha_M \\ \alpha_U \end{bmatrix} \qquad (15)$$

where $L_M, \alpha_M$ are the sets of known pixels, $\alpha_U, L_U$ are the sets of unknown pixels, that is, the area marked foreground and background in Trimap is ranked in the front, and the fuzzy area is ranked in the back. Substitute in equation (13)

$$\begin{aligned} J(\alpha) &= \begin{bmatrix} \alpha_M^T & \alpha_U^T \end{bmatrix} \begin{bmatrix} L_M & B \\ B^T & L_U \end{bmatrix} \begin{bmatrix} \alpha_M \\ \alpha_U \end{bmatrix} \\ &= \alpha_M^T L_M \alpha_M + \alpha_U^T B^T \alpha_M + \alpha_M^T B \alpha_U + \alpha_U^T L_U \alpha_U \end{aligned} \qquad (16)$$

Since $\alpha_M$ is known, we can take the derivative of $\alpha_U$. Let its derivative be 0, and the solution is $\alpha_U = -L_U^{-1}B^T \alpha_M$. It is the optimal solution of the alpha matrix. The value of the alpha matrix is the fusion degree of the foreground. Therefore, we define the locations with values greater than 0.5 in the alpha matrix as weld areas.

### IV. GUIDELINES FOR GRAPHICS PREPARATION AND SUBMISSION

We validate our method on the weld segmentation datasets to test the effectiveness of our segmentation network and edge optimization model. The image data used in the experiments are all from the actual automotive welding production process. Each weld image is obtained by the following steps (1) destructive cutting of the weld area of the automotive, (2) polishing after special chemical etching (3) photographing and then marking the weld area by an expert. So the data set is very difficult to obtain. There are 213 original weld images with resolution of 1024 × 1365, 175 training images, 10 verification images, and 18 test images. Due to the small amount of image data, this paper introduces the data augmentation module to amplify the training data. It is worth noting that there is no data augmentation for the test set and verification set. The segmentation model proposed in this paper is built by PyTorch-1.8 architecture and trained by an NVIDIA Titan V.

### 4.1 Data Augmentation

Data augmentation is widely used in iris recognition, medical image processing, image classification, and other computer vision tasks. It is an effective method to improve model generalization with few training images. Commonly used data augmentation methods can be divided into two ways: topological method and changing image attributes. Topological methods mainly include zooming in and out, figure flipping, center clipping, up and down rotation, perspective deformation, etc. The image attributes vary from brightness, contrast, sharpness, noise, channel, color, etc. This paper uses the Albumentations library provided by Python to augment the training set. Considering the limited GPU memory, data augmentation, and batch reading are carried out simultaneously during the training of the semantic segmentation model. This operation can not only obtain more random data and improve the generalization ability of the model but also avoid excessive consumption of video memory. The flow of the data augmentation module is shown in Fig 8.

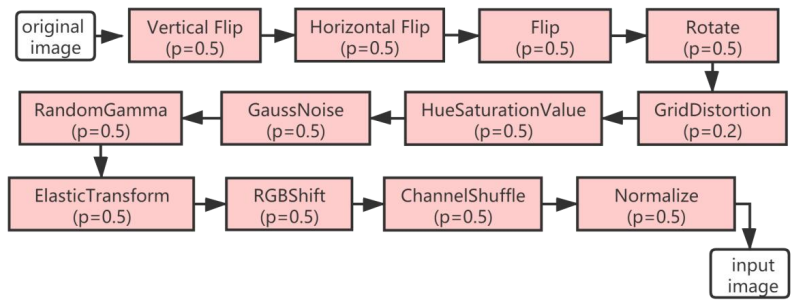

**Fig. 8.** Flow chart of generative random data augmentation module: Where P is the probability of activation of the module, if not activated, enter the next module.

### 4.2 Experimental details

For the semantic segmentation part, we used a pre-trained ResNet-50 as the backbone of the segmentation network, using an SGD optimization algorithm with a momentum of 0.9 and a weight decay factor of 1e-4. In addition, in the early 200 iterative training, we sample the weld image size to 1 / 4 of the original resolution and then input them into the network before training. The learning rate is set to 0.02, and the batch size is set to 8 for rapid training of the segmentation network. In the later stage, 1000 iterative training are carried out with the original image resolution. The learning rate is 0.001, and the batch size is 2. In the edge optimization part of the Matting algorithm, for the hyperparameters $C_{high}$, $C_{low}$ in equation (9), we use a grid search on the training set to find the value that makes MIOU optimal. Final $C_{high}$=0.46, $C_{low}$=0.38.

### 4.3 Experimental results and ablation experiment

This section shows the results of our method on the weld segmentation dataset and compares it with other popular segmentation algorithms. As shown in Table 1, we present the results of the proposed method on the test set. Compared to other segmentation algorithms, our algorithm achieves the best results with 97.93% MIOU. Furthermore, the annotation of the semantic segmentation dataset is subjective, and we additionally invited other experts (Expert II) to annotate the weld regions on the weld dataset. The welds segmentation MIOU of Expert II reached 97.96% based on the original label (Expert I). Thus, it can be found that our algorithm is a relatively accurate segmentation method with the same level of segmentation as that of human experts. Finally, the segmentation effect achieved by our algorithm is shown in Fig 9.

TABLE I

The orange part is the traditional segmentation algorithm, the white part is the deep learning segmentation algorithm, andthe blue part is the manual segmentation

| Mode | Parameters | Mean IOU(%) |
|---|---|---|
| OTSU | no | 48.51 |
| Canny | no | 53.15 |
| WaterShed | no | 56.39 |
| VGG19-UNet | 25,570,786 | 79.38 |
| ResNet50-UNet | 36,535,915 | 82.17 |
| ResNet50-UNet++ | 35,560,843 | 85.63 |
| ResNet34-BASNet | 65,046,112 | 89.68 |
| ResNet50-deeplabv3+ | 39.756962 | 92.67 |
| ResNet50-SHDM-NET | 39,763,616 | 97.93 |
| Human | no | 97.96 |

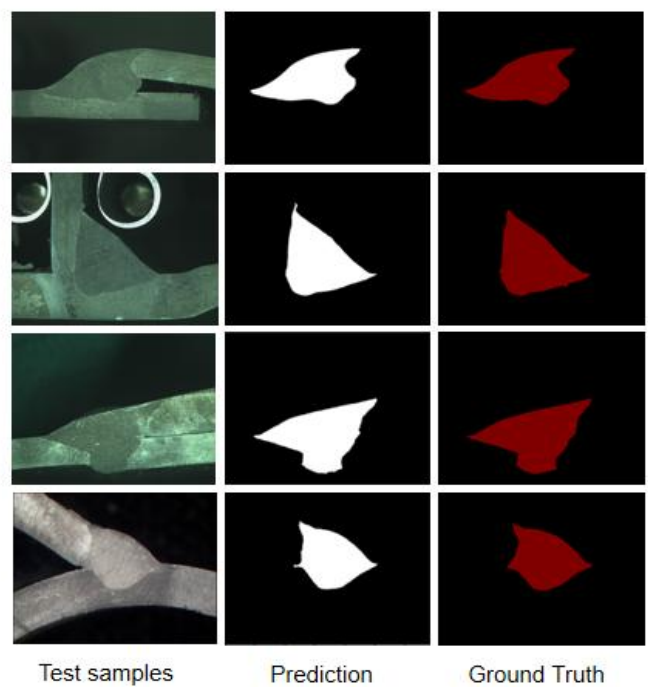

Test samples     Prediction     Ground Truth

**Fig .9.** Visualization results on the test set.

To demonstrate the effectiveness of each part of our proposed segmentation model, we will perform an experimental ablation analysis on each part. As shown in Table 2, the segmentation performance improved from 92.67 to 93.31% after adopting the new semantic segmentation head. We compared and tested the difference in segmentation performance on the welded dataset with and without the heatmap detail guidance module. The Gaussian heatmap guidance module improved the segmentation performance from 93.31% to 97.61%. In order to verify the edge optimization of the Matting algorithm, we tested the performance of the segmentation network with the addition of Matting, as shown in Table (2), and the segmentation performance increased from 97.61% to 97.93% after adding.

TABLE II

"SHDM(pure)" is resnet50+deeplabv3+. "NH" is new semantic segmentation head. "HG" is heatmap detail guidance. All experiments are conducted with the same settings.

| Method | NH | HG | Matting | Mean IOU(%) |
|---|---|---|---|---|
| SHDM(pure) | | | | 92.67 |
| SHDM | ✔ | | | 93.31 |
| SHDM | ✔ | ✔ | | 97.61 |
| SHDM | ✔ | ✔ | ✔ | 97.93 |

V. CONCLUSION

The quality assessment of welded areas of steel plates is based on the segmentation of the welding area. Due to the wide variety of weld regions and the significant difference in noise and color changes, it is difficult for traditional methods to provide a generalized solution. This paper proposes an industrial weld segmentation network based on a deep learning semantic segmentation algorithm fused with heatmap detail guidance and Image Matting. The heatmap guidance module can make the low-level layer pay attention to the details of the segmentation edge, retain the spatial information of the segmentation edge and improve the segmentation performance of the semantic segmentation network. In addition, the Closed-form Matting algorithm is introduced to solve the pain point of the imprecise segmentation edge of the semantic segmentation network, and MIOU reaches 97.93% in the test set.